# milearn: A Python Package for Multi-Instance Machine Learning

D. Zankov[1], P. Polishchuk[2], M. Sobieraj[2], M. Barbatti[1,3]

[1]*Aix Marseille University, CNRS, ICR, Marseille, France*
[2]*Institute of Molecular and Translational Medicine, Palacky University, Olomouc, Czech Republic*
[3] *Institut Universitaire de France, 75231 Paris, France*

**Abstract**
We introduce *milearn*, a Python package for multi-instance learning (MIL) that follows the familiar scikit-learn interface (fit/predict) while providing a unified framework for both classical and neural network–based MIL algorithms for regression and classification. The package also includes built-in hyperparameter optimization designed specifically for small MIL datasets, enabling robust model selection under data-scarce scenarios. We demonstrate the versatility of *milearn* across a broad range of synthetic MIL benchmark datasets: MNIST digit classification and regression, molecular property prediction, and protein-protein interaction (PPI) prediction. Emphasis is placed on the key instance detection (KID) problem, for which the package offers dedicated support.

**Keywords:** multi-instance machine learning, Python

**Source code:** https://github.com/KagakuAI/milearn

## 1. Introduction

Machine learning methods traditionally rely on one-to-one mapping between objects and labels. However, many real-world problems cannot be expressed in this framework, if an object is represented by multiple instances, and it is hard or impossible to select an instance actually linked to a label. In Multi-Instance Learning (MIL) [1], objects are naturally represented by bags of instances, with labels assigned at the bag level rather than to individual instances. This paradigm arises in diverse domains such as computer vision (images represented by collections of patches) [2], natural language processing (documents represented by sets of sentences or paragraphs) [3], remote sensing (satellite images represented by multiple spectral bands or regions) [4], medical diagnostics (whole-slide pathology images represented by collections of tissue tiles) [5], cheminformatics (molecules represented by multiple conformers or fragments) [6], and bioinformatics (protein-protein interactions represented by multiple possible domain pairs) [7].

In many multi-instance scenarios, only a subset of instances within a bag is responsible for the observed label of the whole bag, while the remaining instances contribute noise or redundancy. Key Instance Detection (KID) [8] addresses this problem by identifying which instances are responsible for the outcome. This capability is particularly valuable in applications such as drug discovery, where only certain molecular conformers drive biological activity [6], or in pathology, where only specific tissue regions indicate disease [9]. By highlighting informative instances, KID enhances the interpretability of MIL models.

Several software packages have been developed to support MIL, but they differ in scope and focus. Deep learning-oriented libraries [10], such as *torchmil* [11], focus on neural MIL architectures and are used primarily in domains like computer vision and representation learning. In contrast, traditional and general-purpose MIL toolkits, such as the *mil* library [12] and *misvm* [13], cover more traditional MIL algorithms. Here, *milearn* is introduced as a unified, extensible MIL framework that bridges these two directions. It implements both classical and neural network-based MIL methods while maintaining compatibility with the familiar scikit-learn [14] API (fit/predict) (**Figure 1**). Additionally, *milearn* provides optional stepwise hyperparameter optimization for small datasets and supports key-instance (instance weight) prediction, enabling interpretable analysis alongside model performance.

```python
from milearn.data.mnist import load_mnist, create_bags_reg
from milearn.preprocessing import BagMinMaxScaler
from sklearn.model_selection import train_test_split
from milearn.network.module.hopt import DEFAULT_PARAM_GRID
from milearn.network.regressor import DynamicPoolingNetworkRegressor

# 1. Create MNIST regression dataset
data, targets = load_mnist()
bags, labels, key = create_bags_reg(data, targets, bag_size=10, num_bags=10000,
                                    bag_agg="mean", random_state=42)

# 2. Train/test split and scale features
x_train, x_test, y_train, y_test, key_train, key_test = train_test_split(bags, labels, key,
                                                                          random_state=42)
scaler = BagMinMaxScaler()
scaler.fit(x_train)
x_train_scaled = scaler.transform(x_train)
x_test_scaled = scaler.transform(x_test)

# 3. Train model
model = DynamicPoolingNetworkRegressor()
model.hopt(x_train_scaled, y_train, # recomended for small datasets only
           param_grid=DEFAULT_PARAM_GRID, verbose=True)
model.fit(x_train_scaled, y_train)

# 4. Get predictions
y_pred = model.predict(x_test_scaled) # predicted labels
w_pred = model.get_instance_weights(x_test_scaled) # predicted instance weights
```

**Figure 1. Example of a training regression model for the MNIST dataset**

In this study, we leverage synthetic benchmark datasets to evaluate the performance of *milearn* methods. Synthetic datasets provide a controlled environment in which key instances are explicitly defined, allowing for a direct assessment of a model's ability to identify truly informative instances within a bag - an essential aspect of the multi-instance learning paradigm.

## 2. Package design

The *milearn* design follows a modular structure that makes it easy to prepare data, apply a variety of MIL algorithms, and quickly test MIL models in practical applications.

**Data.** *milearn* introduces a unified representation for bags of instances, where each bag may consist of a variable number of feature vectors and is associated with a single label. Data loaders and preprocessing utilities simplify converting raw datasets into this format. Standard preprocessing steps, such as feature scaling and normalization, can be applied at the bag level.

**Algorithms.** The main module of *milearn* consists of a set of MIL algorithms for both classification and regression. These include classical MIL approaches (such as instance-level and bag-level methods) and neural network-based models. All estimators are implemented with the scikit-learn [14] API (fit, predict), making them directly usable in standard machine learning pipelines.

**Hyperparameters**. To support model selection, *milearn* integrates a hyperparameter optimization module adapted to the MIL setting. It offers stepwise optimization, enabling hyperparameters to be tuned sequentially for improved performance. Because hyperparameter optimization can be computationally intensive, this approach is particularly useful for smaller datasets, where the cost remains manageable while still providing meaningful gains in predictive accuracy.

**Tutorials.** To facilitate adoption, *milearn* is distributed with a collection of tutorial notebooks demonstrating its applications.

## 3. Experiments

To illustrate the versatility of *milearn*, we demonstrate its application across a range of domains, from computer vision to life sciences. Each case highlights different aspects of multi-instance learning and the capabilities of key instance detection. In this study, to perform experiments, besides *milearn*, other complementary packages were involved (**Table 1**), highlighting the maintainability and usability of *milearn*.

**Table 1. Python packages integrated with *milearn* and used in this study**

| Package  | Description                    | Link                                      |
| -------- | ------------------------------ | ----------------------------------------- |
| milearn  | Python MIL                     | https://github.com/KagakuAI/milearn       |
| QSARmil  | Molecular MIL                  | https://github.com/KagakuAI/QSARmil       |
| SEQmil   | Biomolecular MIL               | https://github.com/KagakuAI/SEQmil        |
| QSARcons | Smart ML model consensus search| https://github.com/KagakuAI/QSARcons      |

For simplicity, all experiments were performed with the same MIL method based on multi-instance learning with dynamic pooling [15]. The corresponding source code and more details for all experiments can be found in the tutorials section on the GitHub page (https://github.com/KagakuAI/milearn).

### 3.1 MNIST classification and regression

The MNIST dataset [16] is a widely used benchmark in computer vision, consisting of handwritten digits labeled from 0 to 9. In a conventional supervised setting, each image is associated with a single label. However, MNIST can be reformulated as a multi-instance learning problem by grouping individual digit images into bags (**Figure 2**). This setup enables evaluating both bag-level prediction as well as key instance detection, where the goal is to identify which digits contribute to the bag label.

In the classification setting, a bag was assigned a positive label if it contained at least one instance of a designated *key digit* (e.g., the digit "3"), and a negative label otherwise. The objective was to predict a bag label and simultaneously identify a key digit within the bag (**Figure 2a**). In the regression setting, each bag label was defined as an average of the digits contained in the bag. The model was therefore required to assign meaningful weights to individual digits, reflecting their contribution to the bag average (**Figure 2b**). For the classification experiments, we generated 10000 bags, each containing five images of digits. The dataset was perfectly balanced, with an equal number of positive bags - those containing the key digit "3" - and negative bags, which contained only digits other than "3". For the regression experiments, we constructed a separate dataset of 10000 bags, again with five digits per bag, where the bag's label was defined as the average value of contained digits. Both datasets were randomly divided into training and test sets at 80/20.

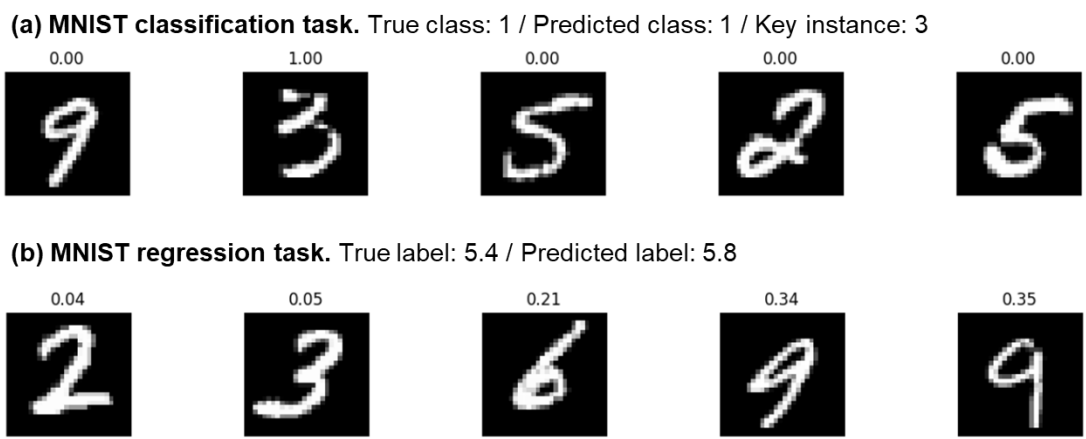

**Figure 2. Example bags with bag label prediction and predicted instance weights for MNIST classification (a) and regression task (b).**

As a result, *milearn* successfully solved both classification and regression tasks. For classification, the model achieved a classification accuracy of 0.96, with KID accuracy reaching 0.99, indicating nearly perfect identification of the key digit. For regression, the model achieved an $R^2$ (coefficient of determination) score of 0.73. KID accuracy for regression was quantified using the correlation of digit ranks: for each bag, the true contribution of each digit to the bag label (its numerical value) and the predicted instance weights were ranked. The KID accuracy was then calculated as the correlation coefficient between original and predicted instance ranks. Using this metric, *milearn* achieved a regression KID accuracy of 0.89, demonstrating its ability to correctly assign higher weights to digits that contribute more to the bag label (**Figure 2b**).

### 3.2 Molecular fragments

Molecules can be viewed not only as whole entities, but also as assemblies of recurring substructures, such as rings, functional groups, or substituents [17]. These fragments influence properties like solubility, hydrophobicity, polarity, and reactivity. Such effects can sometimes be treated as additive, meaning that the overall property of the molecule can be approximated as the sum of the contributions made by its component fragments. *QSARmil* (https://github.com/KagakuAI/QSARmil) is a Python package for constructing multi-instance representations of molecules. Instead of representing each molecule as a single instance, the tool encodes it as a "bag" of instances, which may correspond to conformers, tautomers, or fragments. The package is designed to work in conjunction with the *milearn*, which applies MIL algorithms to the generated multi-instance datasets for classification, regression, and key instance detection.

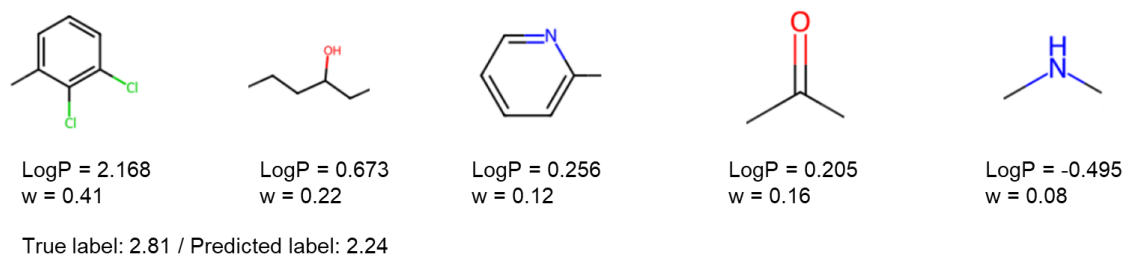

**Figure 3. An example of a bag from the test set containing five fragments with predicted fragment weights.**

In this experiment, bags were represented as collections of substructural fragments (**Figure 3**), where each fragment contributes additively to the overall bag property. For that, each original molecule was fragmented using the BRICS method [17] as implemented in the RDKit package [18], which breaks molecules at specific bonds likely to correspond to synthetically meaningful connection points. As the target property, we used LogP, the logarithm of the partition coefficient between octanol and water, which reflects the hydrophobicity of a molecule/fragment [19]. For each molecule, the bag label was defined as the sum of the LogP values of its fragments, and the objective of the MIL model was to recover this bag-level value while also identifying which fragments contributed most strongly (**Figure 3**). It is important to acknowledge that, although lipophilicity can be regarded to some degree as an additive property and group contribution methods have been developed to estimate the lipophilicity of chemical compounds [20], the lipophilicity of a molecule fundamentally depends on the mutual influence of atoms and fragments in a molecule and thus depends on the chemical context of fragments. In the present study, lipophilicity was treated as an additive property solely for the sake of simplicity.

For this experiment, a dataset of 5012 molecules from the ChEMBL [21] database subsample [6] was divided into a training and test set at 80/20. On the test set, the MIL model achieved an $R^2$ score of 0.85 for bag-level LogP prediction and a KID accuracy of 0.88, calculated as a correlation coefficient between true and predicted contribution ranks of fragments.

### 3.3 Protein–protein interactions

Proteins often work by binding to other proteins, forming what are known as protein-protein interactions (PPIs) [22]. These interactions are usually not driven by the entire protein surface, but instead by small, specific regions - short sequence motifs or structural domains. This problem naturally fits the MIL setting, where a pair of proteins is represented by a set of concatenated protein subsequences (candidate interaction domain pairs).

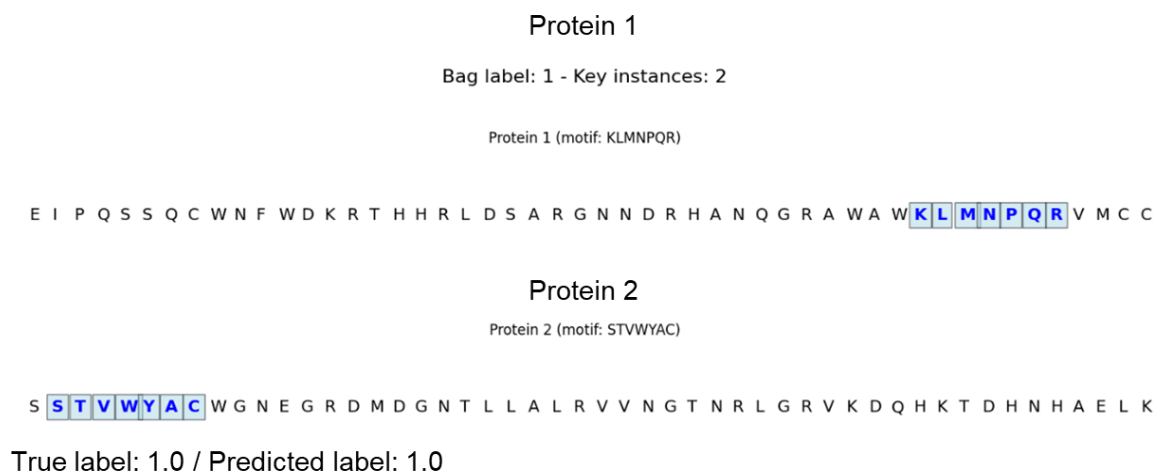

**Figure 4. A pair of proteins with two motifs responsible for positive interactions. The model correctly predicts the highest weight for protein subsequences containing key motifs.**

Therefore, the task for the MIL model is to predict which pairs of proteins interact and which subsequences of proteins are responsible for the interaction. *SEQmil* is a package for constructing multi-instance representations of biomolecular sequences. Instead of treating a protein or nucleic acid sequence as a single entity, *SEQmil* decomposes it into a set of subsequences generated by different segmentation strategies [7]. Each subsequence can then be encoded using a variety of featurization methods. The resulting multi-instance datasets are compatible with standard MIL algorithms - such as those implemented in *milearn*.

To demonstrate the applicability of *milearn* to PPIs modelling, a synthetic dataset was prepared. For each bag, two protein sequences of length 50 were generated randomly. A pair of proteins was labeled positive if both proteins contained the predefined interaction motifs (KLMNPQR and STVWYAC), inserted manually for positive bags (**Figure 4**), and negative otherwise. Each bag was constructed by taking all pairwise combinations of subsequences of length 10 from the two proteins. The key instance in a positive bag corresponded to the subsequence pair that contains both interacting motifs (**Figure 4**). In total, 10000 bags were generated with an equal amount of positive and negative, and were divided into a training and test set at 80/20.

As a result, the MIL model achieved a classification accuracy of 0.99, demonstrating strong performance in predicting whether protein pairs contained interacting motifs. Key instance detection was similarly effective, with a KID accuracy of 1.0, indicating that the model correctly identified at least one pair of subsequences responsible for positive interactions in all test set cases. These results highlight the *milearn* ability to provide both accurate bag-level predictions and interpretable instance-level insights in sequence-based bioinformatics applications.

### 3.4 Hyperparameter optimization and consensus prediction

Hyperparameter tuning is a critical step in machine learning model development, and different approaches to this problem exist, such as grid search, random search, or advanced optimization frameworks [23], [24]. Stepwise hyperparameter optimization, as implemented in *milearn*, offers a practical alternative by tuning parameters sequentially. This approach reduces computational cost while still achieving meaningful improvements in model accuracy. In addition to individual model optimization, consensus prediction provides a robust strategy to improve overall predictive performance by combining predictions from multiple models. Both stepwise hyperparameter optimization and consensus prediction were applied to improve MIL models' performance in the molecular bioactivity prediction task.

MoleculeACE is a benchmark [25] framework designed to evaluate molecular machine learning models with a particular emphasis on activity cliffs - cases where small structural changes in molecules lead to large activity differences. In this work, we use a single dataset (*CHEMBL2147_Ki*) from the MoleculeACE collection. Each molecule was represented as a bag of conformers encoded with 3D descriptors.

**Table 2. Comparison of the single best model and genetic consensus prediction for the ACE benchmark**

|  | Determination coefficient ($R^2$) for the test set | |
|---|---|---|
|  | Default hyperparameters | Optimized hyperparameters |
| Best single model | 0.51 | 0.52 |
| Genetic consensus model | 0.53 | 0.57 |

Nine 3D molecular descriptors were combined with eight MIL algorithms, yielding 72 individual models. The *QSARcons* framework (https://github.com/KagakuAI/QSARcons) was then applied to perform consensus modeling using a genetic search for an optimal combination of built MIL models. The consensus bioactivity predictions were compared with those of the single best-performing model (**Table 2**). In addition, the same modeling pipeline was repeated using the stepwise hyperparameter optimization module from *milearn* (**Table 2**). The results demonstrate that both stepwise hyperparameter optimization and consensus prediction improved the predictive performance of MIL models.

## 4. Conclusion

The *milearn* package provides a unified and flexible framework for multi-instance learning in Python, integrating traditional and modern MIL algorithms, key instance detection, and stepwise hyperparameter optimization within an interface consistent with the scikit-learn ecosystem. Through experiments on a diverse set of benchmarks, *milearn* demonstrates its ability to support both accurate bag-level prediction and interpretable instance-level insight across diverse application domains.